\documentclass[runningheads]{llncs}
\usepackage[T1]{fontenc}
\usepackage{graphicx}
\usepackage{amsmath}
\usepackage{bbding}
\usepackage{pifont}
\usepackage{wasysym}
\usepackage{amssymb}
\usepackage{multirow}
\usepackage{color}
\usepackage{hyperref} 
\begin{document}
%
\title{Total Disentanglement of Font Images into Style and Character Class Features}
\titlerunning{Total Disentanglement of Font Images ...}
%
\author{Daichi Haraguchi\inst{1,2}\and
Wataru Shimoda\inst{2} \and
Kota Yamaguchi\inst{2} \and
Seiichi Uchida\inst{1}}
\authorrunning{D. Haraguchi et al.}
%
\institute{Kyushu University, Japan \and CyberAgent, Japan \\
\email{haraguchi\_daichi\_xa@cyberagent.co.jp}\\}

\maketitle              
\begin{abstract}
In this paper, we demonstrate a total disentanglement of font images. 
Total disentanglement is a neural network-based method for completely decomposing each font image into its style and content (i.e., character class) features. 
It uses a simple but careful training procedure to extract the common style feature from all `A'-`Z' images in the same font and the common content feature from all `A' (or another class) images in different fonts. 
These disentangled features guarantee the reconstruction of the original font image.
Various experiments have been conducted to understand the performance of total disentanglement. First, it is demonstrated that total disentanglement is achievable with very high accuracy; this is experimental proof of the long-standing open question, ``Does `A' exist?'' Hofstadter (1985). 
Second, it is demonstrated that the disentangled features produced by total disentanglement apply to a variety of tasks, including font recognition, character recognition, and one-shot font image generation.
Code is available here: \url{https://github.com/uchidalab/total_disentanglement}
%

\keywords{Font analysis  \and Disentanglement \and Font generation.}
\end{abstract}
%
%
\section{Introduction}\label{sec:intro}
\begin{figure}[t]
    \centering
    \includegraphics[width=0.65\textwidth]{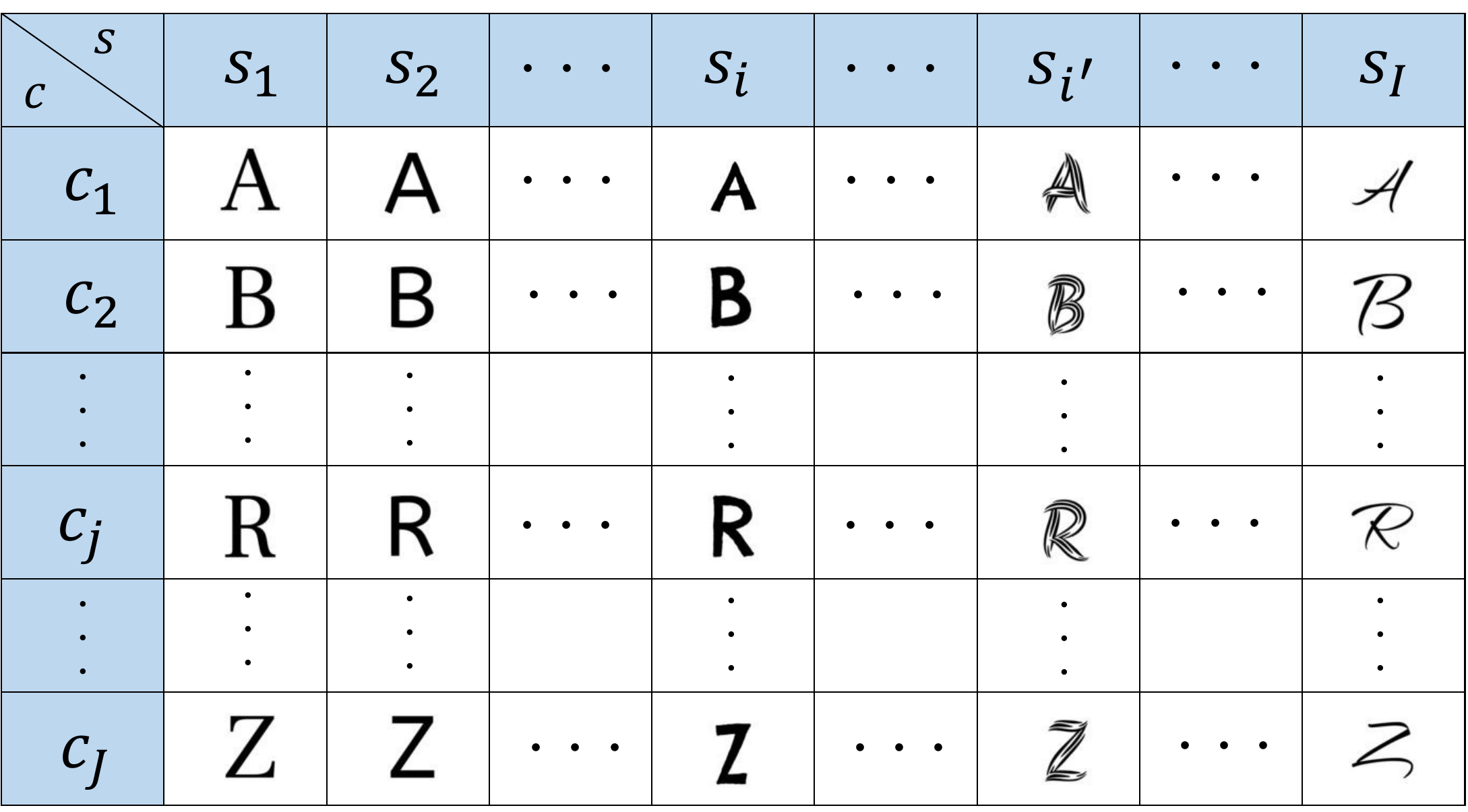}\\[-3mm]
    \caption{A matrix of $I$ font styles and $J$ characters. Our goal is to extract an (almost) identical style feature from each column $i$ and an (almost) identical content (character class) feature from each row $j$.}
    \label{fig:matrix}
\end{figure}

The letter `A' appears in various styles depending on its font.
Fig.~\ref{fig:matrix} shows Latin alphabet letters $c_j\in \{\text{`A,'}...,\text{\rm `Z'}\}$ in $I$ different fonts, $s_1, \ldots, s_I$.
When `A' is printed with a \textit{serif} font (like $s_1$), serifs appear at its stroke endpoints; when \textit{sans-serif} (like $s_2$), no serif appears.
Aside from serifs, there are numerous other design elements to consider, such as corner
shapes, stroke thickness, curves, and decorations. Interestingly, we can recognize all of these `A's as `A,' regardless of their style or whether they are new to us.
\par
In 1985, Hofstadter who is a cognitive scientist and a philosopher asked a question ``{\it Do all `A's have a common `A'-ness?}\ '' in his famous book, {\it Metamagical Themas}~\cite{metamagical1985}.
He also asked ``{\it Do all characters printed in a font have a common spirit (i.e., style)?}\ ''
Finally, he conjectured that any character image should be divided into `x'-ness and style, 
and a pair of `x'-ness and style compose a character image in its style.
He defined this conjecture as \textit{``the vertical and horizontal problem''}, 
by showing vertically and horizontally arranged characters like Fig.~\ref{fig:matrix}.
\par
\begin{figure}[t]
    \centering
     \includegraphics[width=1.0\textwidth]{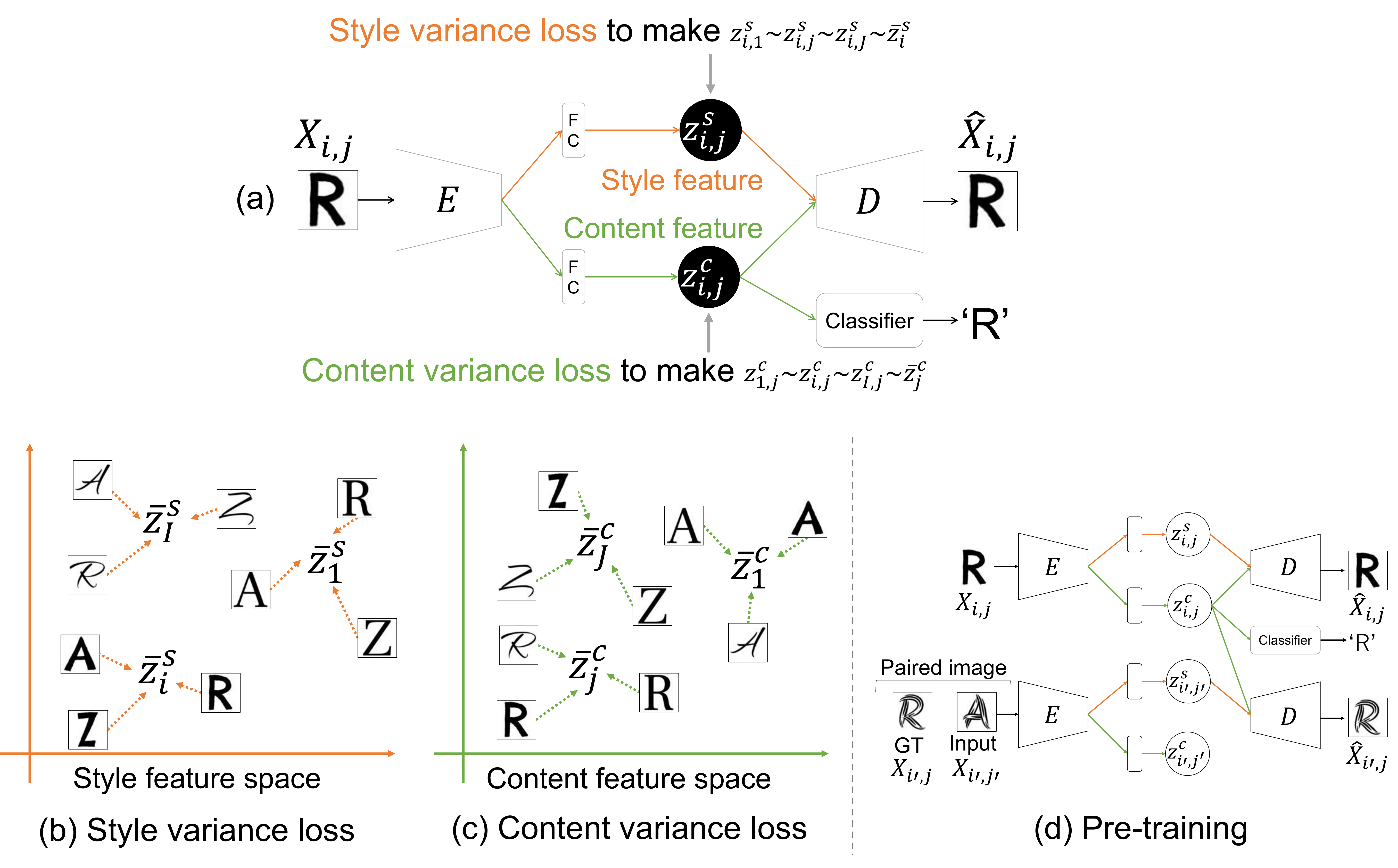}\\[-3mm]
 \caption{(a)~The overview of the proposed method, total disentanglement. $E$ and $D$ indicate an encoder and decoder, respectively.  (b)~Style variance loss and (c)~content variance loss. The distance between an average style feature $\bar{z}^s_i$ and each style feature in a font $s_i$ is minimized by (b). Similarly, the distance between an average content feature $\bar{z}^c_i$ and each style feature in character $c_j$ is minimized by (c). (d)~Overview of our pretraining approach. See for detail of the pretraining in Section~\ref{sec:stabilize}.\label{fig:overview}}
\end{figure}

In this study, we propose \textit{total disentanglement} to experimentally prove that Hofstadter's vertical and horizontal problem is solvable. 
Disentanglement is a common approach to separating latent factors---such as style and content---into independent, interpretable components.
Fig.~\ref{fig:overview} (a) shows an overview of the proposed method, which decomposes an arbitrary font image into its font style feature (hereinafter referred to as a {\em style feature})  and character class feature (hereinafter, a {\em content feature}). We expect the style feature to be common in the $i$th font (e.g., $i$th column in Fig.~\ref{fig:matrix}), and the content feature of the $j$th character class to be common (e.g., $j$th column in Fig.~\ref{fig:matrix}). 
Furthermore, the decomposition is near-lossless; we can almost completely reconstruct the input font image from the style and content features.

The decomposed style and content features are purely representing their style and class; therefore, they are useful for several applications. 
For example, the content feature will be useful for font-independent character recognition; similarly, the style feature is useful for character-independent font recognition. 
A more important application is {\em one-shot} font generation; by having style features of a font image with $s_i$ and then combining it with content features of $c_j$, it is possible to generate all the other images with $s_i$. 
In the subsequent experiments, we will demonstrate that total disentanglement performs well in those applications.
\par
The proposed method, as shown in Fig.~\ref{fig:overview}~(a), employs a simple but careful training process that takes advantage of the unique property of fonts that font images can be arranged in a matrix, as shown in Fig.~\ref{fig:matrix}. Total disentanglement has an encoder and a decoder like an autoencoder, and its latent variable is divided into style and content features. The use of two loss functions is a technical highlight. 
One is called style variance loss which forces all style features from $s_i$ to be converged to their average style features, as shown in  Fig.~\ref{fig:overview}~(b). The other is called content variance loss which forces all content features from $c_j$ can be converged to their average content features. 
Also, note that these variance losses only apply to samples that can be arranged as a matrix, as shown in Fig.~\ref{fig:matrix}. These concepts distinguish total disentanglement from existing disentanglement methods, which are discussed in the following section.
\par 
To avoid trivial solutions, we also include an elaborated pre-training step. Directly minimizing the two variance losses leads to a trivial solution in which all feature vectors become the zero vector or a specific common vector. (Recent contrastive learning methods that employ some distance minimization losses~\cite{hua2021feature,chen2021exploring} also encounter this trivial solution, called ``collapsing.'') 
As we will see later, the style transfer framework is used as a pre-training step to initialize the average style vectors of $s_i$ and average content vector of $c_j$, as well as the encoder and decoder models.
\par

We summarize our contributions as follows.
\begin{itemize}
    \item We propose the total disentanglement method for font images. The proposed method can disentangle a font image into two components: a content feature, common to all images in a character class, and a style feature, common to all images in a font set.
    \item We experimentally demonstrate that the disentangled features have the expected properties and thus perform better than competitive methods in various tasks, such as font recognition, character recognition, and one-shot font generation.
    \item To the best of the authors' knowledge, this is the first proof that Hofstadter's vertical and horizontal problems can be solved using nonlinear machine learning experiments. At the same time, these positive results indicate the existence of `A'-ness (, or `x'-ness) as the Platonic idea of `A.'
\end{itemize}

\section{Related work}\label{sec:related-work}
\subsection{Feature disentanglement}\label{sec:related-work-disentangle}
Disentanglement is a technique that uses neural networks to decompose an input sample into two independent features. For example, a face image can be disentangled into its person-specific and emotion-related features using an adversarial training strategy~\cite{liu2018exploring}. This method is directly applicable to font images by simply consider person-specific features as character class features and emotion-related features as font style features, as shown in Fig.~\ref{fig:comparative_model}~(b). 

\par
Disentanglement can be used for a style transformation. For example, it decomposes an image into the feature representing the content objects and the style feature~\cite{gonzalez2018image}; and then visualizes the object in a different style by modifying or replacing the style feature. By a similar mechanism, DRIT++~\cite{lee2020drit++} can convert photographic images into paintings. 
Domain adaptation is another application of disentanglement~\cite{lee2021dranet,cai2019learning}. Individual samples can be disentangled into their domain-independent features and 
domain-dependent features, and domain adaptation can be realized by modifying the latter. \par
%

\subsection{Disentanglement for font images}\label{sec:font-disentangle}
\begin{figure}[t]
    \centering
    \includegraphics[width=0.75\textwidth]{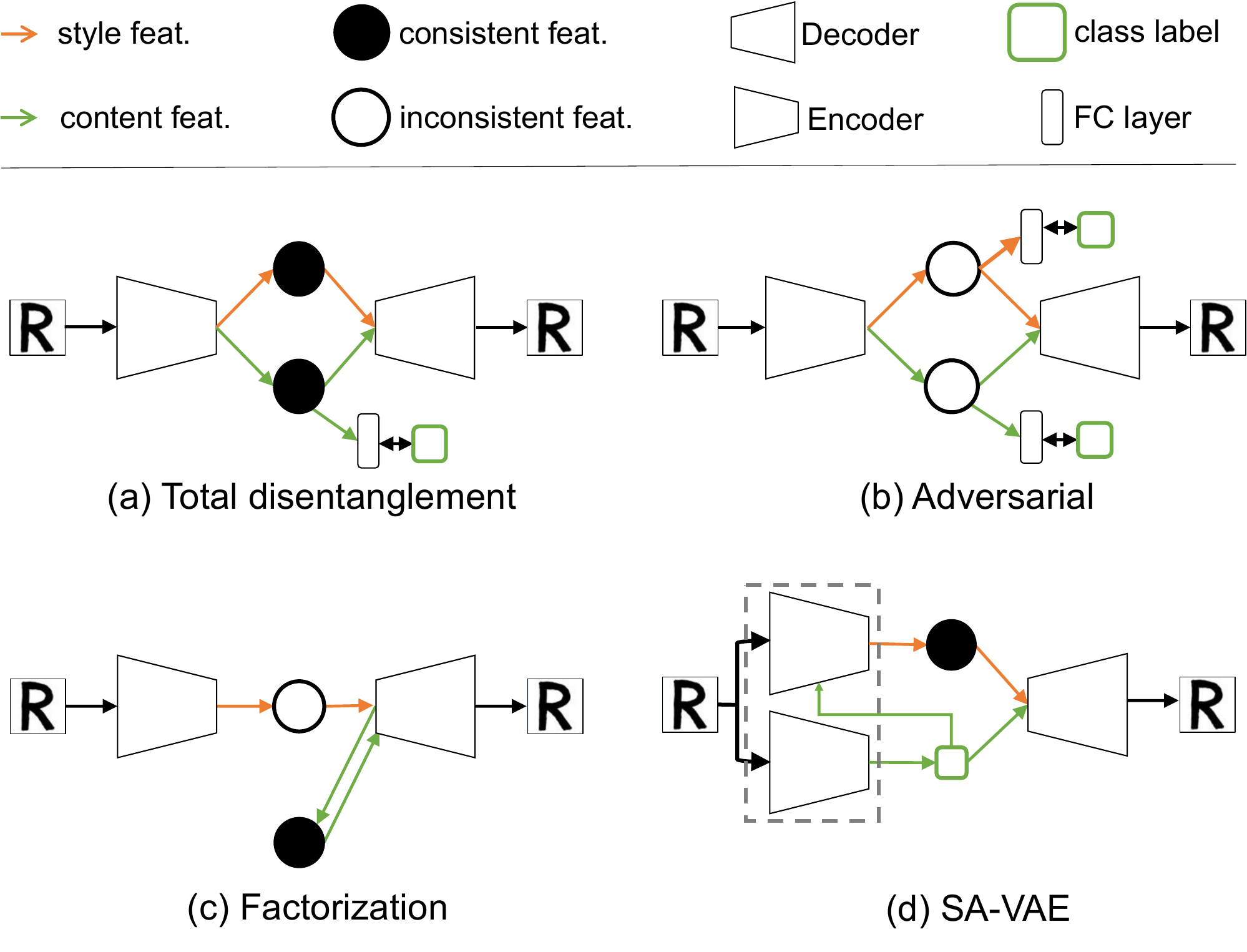}\\[-3mm]
    \caption{Various disentanglement methods for font images. 
     Orange and green arrows show the style features and content features, respectively.}
    \label{fig:comparative_model}
\end{figure}

Disentanglement has recently been applied to font images in order to extract character-class independent style features. Disentanglement is mostly used for few-shot or one-shot font generation tasks~\cite{sun2018learning,srivatsan2019deep,cha2020few,srivatsan2021scalable,gao2019artistic,yang2019tet,yang2020te141k,yang2024fontdiffuser,he2024diff,fu2024generate}. 
One-shot or few-shot font generation is a common approach, wherein only one or a few examples (e.g., `A' and `B') of a target font are provided, and these styles are then used to generate all the remaining characters (`C' to `Z'). Most approaches aim to achieve font generation by focusing on font style disentanglement, while the content is not disentangled. This is due to the difficulty of extracting precise content features from character images across a variety of font styles. Instead, content features are extracted from a specific font~\cite{yang2024fontdiffuser}, or content labels are directly provided~\cite{roy2020stefann}. 
In another example, Srivatsan~\textit{et al.}~\cite{srivatsan2019deep} disentangle font style features using a variational autoencoder-based model and utilize trainable parameters as content features, as shown in Fig.~\ref{fig:comparative_model} (c). A few approaches attempt to extract not only font style features but also content features~\cite{sun2018learning,zhang2018separating}, as shown in Fig.~\ref{fig:comparative_model} (d).
\par

In contrast to conventional disentanglement approaches for font images, our approach focuses not on font generation but on precisely disentangling both font style features and character content features.
Thanks to the total disentanglement of font images, our approach enables one-shot font generation by combining the disentangled features.


\par

\section{Total disentanglement of character class and font style}
\subsection{Overview\label{sec:technical-overview}}
Hofstadter's vertical and horizontal problem, which we described in Section~\ref{sec:intro}, can be formally explained as follows.
Let $X_{i,j}$ denote a character image of a character class $c_j\in \{\text{`A,'}\ldots,\text{\rm `Z'}\}$\footnote{Throughout this paper, we focus on the Latin alphabet, which is comprised of 26 capital letters.
In theory, our method is applicable to other alphabets, including Asian language alphabets. We used the Latin alphabet because we can easily collect many alphabets in various styles.} printed in a font style $s_i\in S$, where $S$ denotes a set of $I$ fonts. Then, the vertical and horizontal problem asks ``{\it Can we decompose  $X_{i,j}$ into two feature vectors, a style feature of $s_i$ and a content (i.e., character class) feature of $c_j$, and reconstruct $X_{i,j}$ from them?}\ ''  Despite its philosophical and computer-scientific significance, few studies have addressed it since Hofstadter defined the problem.

To solve the above problem, we propose an autoencoder that can disentangle a font style feature and a content feature from a font image. 
As shown in Fig.~\ref{fig:overview}~(a), an autoencoder is used for disentangling style and content features. Specifically, a font image $X_{i,j}$ is fed into an encoder $E$, and then its output is further fed into two fully connected layers for obtaining a style feature $z^s_{i,j}$ and a content feature $z^c_{i,j}$. 
A decoder $D$ outputs a reconstructed image $\hat{X}_{i,j}$ from $z^s_{i,j}$ and $z^c_{i,j}$. 
\par
The proposed model is trained by the two variance loss functions detailed in Section~\ref{sec:variance-loss} and two other loss functions detailed in Section~\ref{sec:other-loss}. 
We use a pre-training step to stabilize the training process before training the model with those loss functions. The pre-training step is detailed in Section~\ref{sec:stabilize}.\par
\subsection{Variance loss\label{sec:variance-loss}}
We propose two variance loss functions, $L_\mathrm{style}$ and $L_\mathrm{content}$, to have disentangled style features and content features that satisfy the conditions $z^s_{i,1}\sim z^s_{i,j}\sim z^s_{i,J}$ for any $i$ and  $z^c_{1,j}\sim z^c_{i,j}\sim z^c_{I,j}$ for any $j$. More specifically, we use 
{\em average} features $\bar{z}^s_i$ and $\bar{z}^c_j$ as the guides to satisfy these conditions, as shown in Figs.~\ref{fig:overview}~(b) and (c). The derivation of the average features will be detailed later. 
Consequently, the variance losses $L_\mathrm{style}$ and $L_\mathrm{content}$ are formulated as
\begin{eqnarray}
   L_\mathrm{style} &=& \frac{1}{IJ}\sum_i^{I}\sum_j^{J}{\|z^s_{i,j} - \bar{z}^s_i \|_2^2}, \label{eq:font_variance}\\
   L_\mathrm{content} &=& \frac{1}{IJ}\sum_i^{I}\sum_j^{J}{\|z^c_{i,j} - \bar{z}^c_j \|_2^2}.\label{eq:char_variance}
\end{eqnarray}
\subsection{Other losses\label{sec:other-loss}}
Since we expect $\hat{X}_{i,j}\sim X_{i,j}$, the mean absolute error (MAE) loss is simply used as the reconstruction loss:
\begin{equation}
    L_\mathrm{rec} = \frac{1}{IJ}\sum_{i}^{I}\sum_{j}^{J} \|X_{i,j} - \hat{X}_{i,j} \|.
\end{equation}
\par
For letting the content feature $z^c_{i,j}$ surely carry the character class information, we employ a classifier $F$ for $z^c_{i,j}$ as shown in Fig.~\ref{fig:overview}~(a), which is trained by the standard cross-entropy loss:
\begin{equation}
    L_\mathrm{cls} = -\frac{1}{IJ}\sum_i^{I}\sum_j^{J}{y_j\log{F(z^c_{i,j})}}, 
\end{equation}
where $y_i$ is the $J$-dimensional one-hot vector showing the character class.
As the classifier $F$, we employ a simple multi-layer perceptron (MLP) with two fully connected layers and the softmax function. 

\subsection{Pre-training step}\label{sec:stabilize}
As noted in Section~\ref{sec:intro}, direct minimization of the two variance losses yields trivial solutions: all feature vectors become zero or a meaningless common vector. Furthermore, we are unaware of the average vectors, $\bar{z}^s_i$ and $\bar{z}^c_j$, which guide the variance minimization of  (\ref{eq:font_variance}) and (\ref{eq:char_variance}). As a result, we introduce a pre-training step, as shown in Fig.~\ref{fig:overview}~(d), before training our main model of Fig.~\ref{fig:overview}~(a). \par
%
In the pre-training step, we do not think of variance minimization; instead, we try to have an encoder $E$ that disentangles $X_{i,j}$ into $z^s_{i,j}$ and $z^c_{i,j}$ and 
a decoder $D$ that reconstructs $\hat{X}_{i,j}\sim X_{i,j}$ from then. Note that we expect neither $z^s_{i,1}\sim z^s_{i,j}\sim z^s_{i,J}$ nor  $z^c_{1,j}\sim z^c_{i,j}\sim z^c_{I,j}$ because no variance losses are used. 
Furthermore, we train $E$ and $D$ with a style transferring framework, where three images $X_{i,j}$, $X_{i',j}$, and $X_{i',j'}$ are used at each training step. More specifically, we expect that the reconstructed image $\hat{X}_{i',j}$ from the content feature $z^c_{i,j}$ from $X_{i,j}$ and the style feature $z^s_{i',j'}$ from $X_{i',j'}$ becomes similar to $X_{i',j}$.
\par
Consequently, we pre-train $E$ and $D$ with the following loss function $ L_\mathrm{pre}$:
\begin{equation}
   L_\mathrm{pre} = L_\mathrm{rec} + L_\mathrm{trans} + \lambda_\mathrm{cls} L_\mathrm{cls},
\end{equation}
where $\lambda_\mathrm{cls}$ is a hyperparameter and $L_\mathrm{trans}$ evaluates the quality of the style-transferred image  $\hat{X}_{i',j}$: 
\begin{equation}
      L_\mathrm{trans} = \frac{1}{IJ}\sum_{i'}^{I}\sum_{j}^{J} \|X_{i',j} - \hat{X}_{i',j}\|.
\end{equation}
\subsection{Fine-tuning}
After pre-training, we fine-tune the model (in Fig.~\ref{fig:overview}~(a)) for the total disentanglement with the variance losses. Before fine-tuning, the average feature $\bar{z}^s_i$ is prepared by averaging $z^s_{i,j}$ for all $i$ and $j$ given by the pre-trained encoder. The other average feature $\bar{z}^c_j$ is prepared similarly. Then the model is fine-tuned with $L_\mathrm{fine}$:
\begin{equation}\label{eq:finetune_loss}
   L_\mathrm{fine} = L_\mathrm{rec}  + \lambda_\mathrm{cls} L_\mathrm{cls} + \lambda_\mathrm{style} L_\mathrm{style} + \lambda_\mathrm{content} L_\mathrm{content},
\end{equation}
where $\lambda_\mathrm{style}$ and $\lambda_\mathrm{content}$ are hyperparameters.\par
%

\section{Experimental Results}\label{sec:experiment}
\subsection{Font datasets}
For training the model, we used Google Fonts\footnote{\url{https://github.com/google/fonts}}, a widely used font database.
Specifically, 2,631 and 315 fonts were used for training and validation, respectively.
An additional 295 fonts were employed as test data for the feature-visualization task.
The paired images required for the pre-training step were randomly selected. 
The following examples focus on 26 capital letters from `A' to `Z,' as in~\cite{srivatsan2019deep,liu2022learning}.\par
To ensure complete separation between training and test samples, we employed 1,132 fonts in Adobe font~\footnote{\url{https://www.adobe.com/products/fontfolio.html}} for evaluation. Since most fonts in the dataset are ``formal,'' we also employed 1,502 fonts in Capitals64~\cite{azadi2018multi} dataset, which contains more fancy fonts.  
Note that we excluded 58 fonts from the original 1,560 test fonts of Capitals64 because the same fonts can be found in Google Fonts.\par
\begin{figure}[t]
     \centering
     \includegraphics[width=0.85\linewidth]{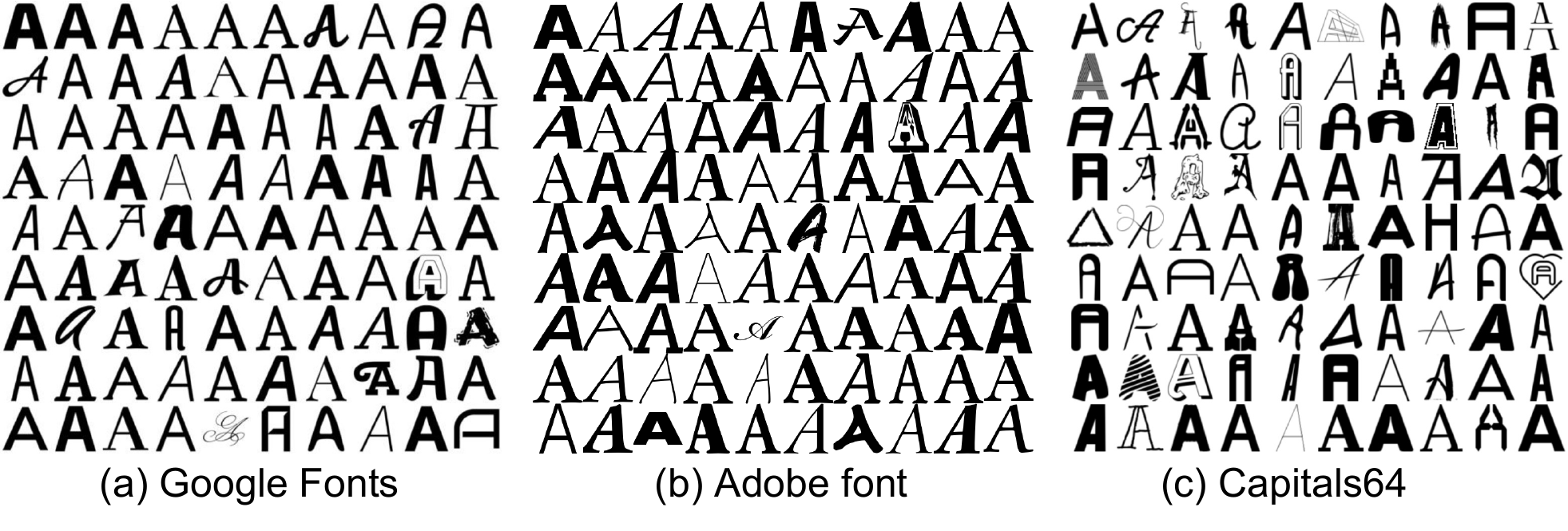}\\[-4mm]
 \caption{Examples of `A' in each dataset.}
    \label{fig:dataset}
\end{figure}

Fig.~\ref{fig:dataset} shows image examples of `A' from each of the three font datasets. They show large shape variations. Especially, Capitals64, in particular, has many fancy fonts. These variations demonstrate how difficult it is for us to extract (almost) the same content feature from all those `A's. Furthermore, we did not use Capitals64 for training. As a result, disentangling unseen fancy fonts in Capitals64 during the testing phase is extremely difficult.

\subsection{Implementation details}~\label{seq:implementation-details}
For the encoder and the decoder, we used simple convolution and deconvolution layers. The encoder consists of four convolution layers, and the decoder reverses the encoder. 
Except for the last deconvolution layer, all convolution layers and deconvolution layers include batch normalization and a ReLU function. 
The last deconvolution layer has a sigmoid function.
Each branch between the encoder and decoder consists of two FC layers with a 128-dim output. Each font style and the content feature can be obtained as 128-dim vectors from this. 
These two features are concatenated and fed into the decoder after upsampling.
Additionally, all FC layers have a ReLU function.
The character classifier consists of an FC layer with a sigmoid function. 
\par
All input images were $64\times64$ gray-scale.
Loss weights were set as $\lambda_\mathrm{cls}=0.001$, $\lambda_\mathrm{style}=1$, and $\lambda_\mathrm{content}=1$, by using validation samples. 
The batch size was 64, and the pre-training and training epochs were 23 and 15, respectively, when early stopping was used.
We employed the Adam optimizer with a learning rate of $0.001$.

\subsection{Comparative methods}\label{sec:comparative-methods}
In the experiment, we used the following six comparative methods.
\textbf{Adversarial} (shown in Fig.~\ref{fig:comparative_model}~(b)) is a disentangle method with a model structure similar to ours, although it is trained with our loss functions used in our pretrained model and two loss functions including adversarial loss for disentanglement described in~\cite{liu2018exploring}.
\textbf{SA-VAE}~\cite{sun2018learning} (Fig.~\ref{fig:comparative_model}~(c)) is the VAE-based disentanglement model. This method provides the content feature as a one-hot vector. 
The official implementation is not available, and the details of the model structure are not described; therefore, we implement the model with a similar structure to ours.
\textbf{Factorization}~\cite{srivatsan2019deep} (Fig.~\ref{fig:comparative_model}~(d)) is a font generation model that focuses more on style feature extraction rather than on being a disentanglement model. This model does not extract content features but instead uses trainable feature vectors as content features.
\textbf{FANnet}~\cite{roy2020stefann} is also font generation model based on a style feature extractor and thus has a similar limitation as Factorization. This model uses the content label when generating font images.
The model structure is similar to Factorization.
Note that Factorization and FANnet can not perform content disentanglement (i.e., can not extract content features).
\textbf{w/o fine-tune} is an ablation model that does not require fine-tuning. In other words, it is a pre-trained model that has not been trained with variance losses. 
\par

 \begin{figure}[t]
    \centering
    \centering
    \includegraphics[width=\linewidth]{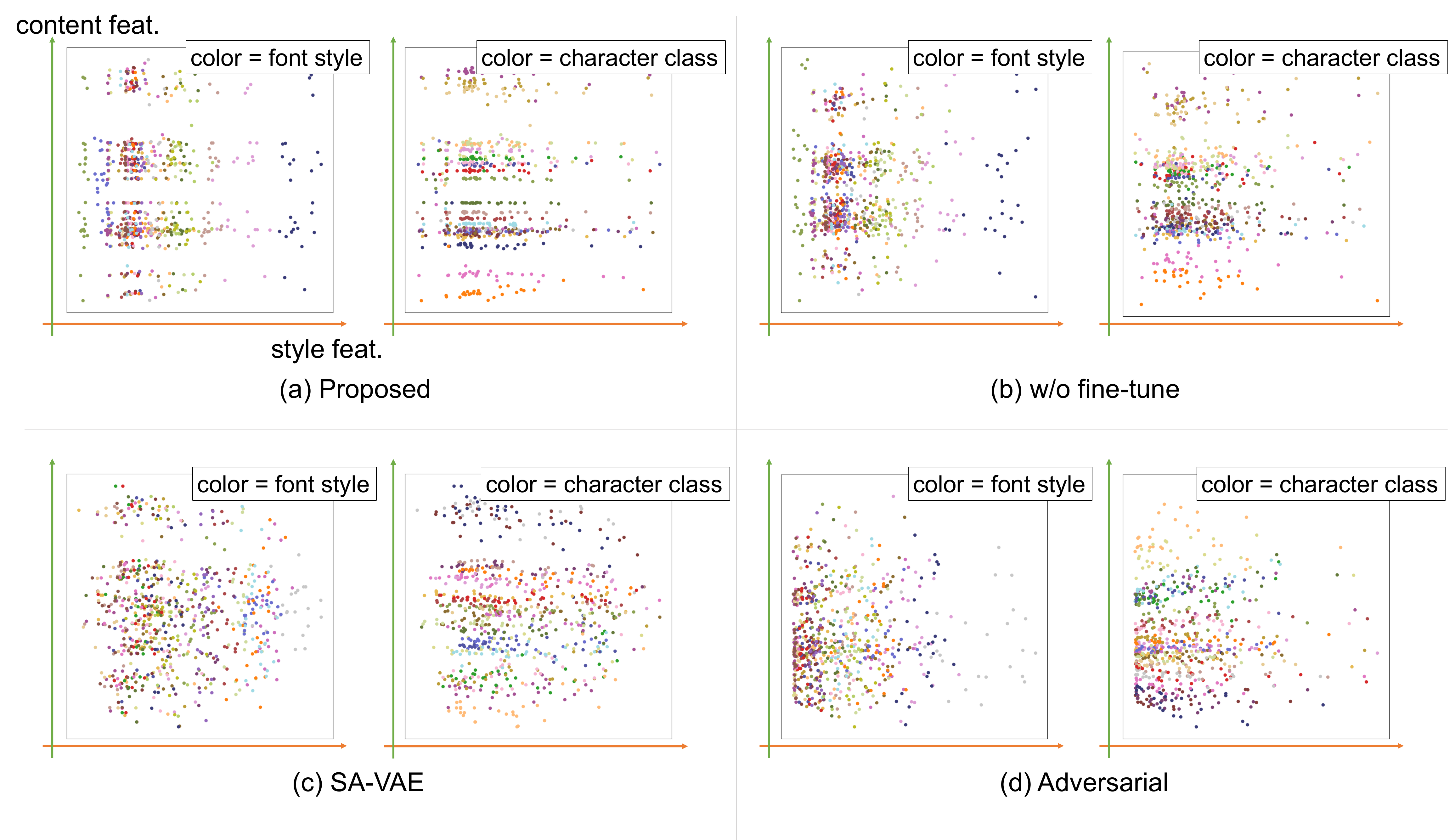}\\[-3mm]
 \caption{Visualization of font style and content features by using PCA in the same space. Except for the coloring rule, the left and the right plots in each subgraph are identical; the left uses the same color for the same font and the right for the same character class.}
    \label{fig:matrix-pca}
\end{figure}
\subsection{Visualization of style features and content features}
Fig.~\ref{fig:matrix-pca} shows the style and content features extracted from test samples from Google Fonts. For this visualization, each of the style feature $z^s_{i,j}$ and the content feature $z^c_{i,j}$ is mapped to be one-dimensional independently. For dimensionality reduction, principal component analysis (PCA) is used. As a result, two axes of Fig.~\ref{fig:matrix-pca} correspond to the first principal components of $\{z^s_{i,j}\}$ and $\{z^c_{i,j}\}$, respectively. 
\par
Fig.~\ref{fig:matrix-pca} reveals two important facts. Firstly, as shown in Fig.~\ref{fig:matrix-pca}~(a), the proposed method could realize the total disentanglement of style and content features, as expected. From the left plot, we can see that the dots in the same color distribute like a vertical straight line --- this result implies that the 26 images (`A'-`Z') could have almost the same style feature. 
From the right plot, the dots in the same color form a horizontal line --- implying that all of `A' (or another character class) have the same content feature. As a result, Fig.~\ref{fig:matrix-pca}~(a) suggests that Hofstadters' ``\textit{the vertical and horizontal problem}''~\cite{metamagical1985} is solvable by our total disentanglement.
\par 
Secondly, by comparing Fig.~\ref{fig:matrix-pca}~(a) with (b), the fine-tuning step is necessary for a more complete disentanglement. 
The colored-line structures in (a) are destroyed in (b) (at least in this one-dimensional + one-dimensional space). In other words, our variance losses must have font-consistent style features as well as character class-consistent content features. 
About the result that all `A's could have mostly the same content features, we will make an extra discussion in Section~\ref{sec:discussion}.\par
Fig.~\ref{fig:matrix-pca}~(c) visualizes the content and style features extracted by SA-VAE~\cite{sun2018learning}. SA-VAE takes care of ``pair-wise'' style consistency between two images, $X_{i,j}$ and $X_{i,j'}$. Consequently, its consistency becomes less tight than ours, where all $X_{i,1},\ldots, X_{i,j},\ldots, X_{i,J}$ are requested to have (almost) the same style feature. This difference makes the feature distribution of SA-VAE less consistent. 
Additionally, on the right of (c), the colored-line trend that can be seen in (a) can not be clear.
In (c), content features are extracted from CNN trained through character classification.
In contrast, the proposed method shows a more clear trend on the right of (a). 
This difference comes from the `A'-ness obtained by the content variance loss.\par

Fig.~\ref{fig:matrix-pca}~(d) visualizes the content and style features extracted using the adversarial approach based on~\cite{liu2018exploring}. This approach effectively disentangles the style and content features. However, explicit control of feature consistency has not been considered. Consequently, the feature distributions are less consistent than ours.
\subsection{Quantitative evaluation of the disentangled features}
\subsubsection{Font recognition} \label{sec:experiment1}
To quantitatively evaluate the disentangled style features, we conduct
a font recognition experiment using test samples. 
We will have a high font recognition accuracy if our style features purely represent font styles.
For this evaluation, we extracted style from test samples (i.e., Adobe font and Capitals64) and then split them for training and testing an MLP-based classifier for font recognition\footnote{Using a different classifier for feature quality evaluation can be found in another literature, such as \cite{sun2018learning}.}.  
Specifically, we randomly split 26 character classes into 13 and 13.
Second, we chose 100 fonts from each Adobe font and Capitals64.
We obtained 1,300 training samples and 1,300 test samples by extracting the style feature from each character.
Third, we trained and tested an MLP (composed of three fully connected layers with a softmax function) for font recognition using those samples (with 100 classes). As noted above, if it achieves higher accuracy, it demonstrates that our style features accurately represent font style.
\par
\begin{table}[t]
 \caption{Test accuracy (\%) of font recognition using style features or content features. Note that Factorization and FANnet can not content disentanglement.}
 \vspace{-2mm}
 \label{table:nn-font}
 \centering
  \begin{tabular}{l|r|r|r|r}
    \hline
         & \multicolumn{2}{c|}{Adobe font} & \multicolumn{2}{c}{Capitals64} \\ \cline{2-5}
         & \small{style feat.\!$\uparrow$} & \small{content feat.\!$\downarrow$} & \small{style feat.\!$\uparrow$} & \small{content feat.\!$\downarrow$}\\ \hline\hline
        Proposed & \textbf{77.57} & \textbf{1.24} & \textbf{70.17} & \textbf{3.83} \\ 
        w/o fine-tune  & \underline{71.27} & 2.60 & \underline{65.45} & 7.32 \\\hline
        Adversarial & 57.63 & 1.78 & 49.25 & 5.64 \\
        SA-VAE~\cite{sun2018learning} & 57.11 & \underline{1.45} & 45.81 & \underline{4.42} \\ 
        Factorization~\cite{srivatsan2019deep} & 37.95 & N/A$^*$ & 30.67 & N/A$^*$ \\ 
        FANnet~\cite{roy2020stefann} & 37.19 & N/A$^*$ & 36.11 & N/A$^*$ \\\hline
        
  \end{tabular}

\end{table}

Table~\ref{table:nn-font} shows the test accuracy of font recognition by the style features. The accuracy values in this table are based on an average of ten trials with different character class splits and 100 font selections. This result showed that the proposed method achieves higher accuracies ($77.57\%$ and $70.17\%)$ than the comparative methods for both datasets. 
Furthermore, the comparison with ``w/o fine-tune'' proves the validity of the variance losses.
\par 
We also performed another evaluation by training a font classifier with the extracted {\em content} features. If there is no information about font style in the content features, the recognition accuracy should be very low. Since there are 100 font classes in this evaluation, the ideal accuracy is 1\%. As shown in Table~\ref{table:nn-font}, among all methods, the proposed method performs the best performance. It almost reaches ideal accuracy, especially for Adobe font.
\begin{table}[t]
 \caption{Test accuracy (\%) of character recognition using content features or style features. Note that Factorization and FANnet can not perform content disentanglement.}
 \vspace{-2mm}
 \label{table:nn-char}
 \centering
  \begin{tabular}{l|r|r|r|r}
    \hline
         & \multicolumn{2}{c|}{Adobe font} & \multicolumn{2}{c}{Capitals64} \\ \cline{2-5}
         & \small{content feat. $\uparrow$} & \small{style feat. $\downarrow$} & \small{content feat. $\uparrow$} &\small{ style feat. $\downarrow$}\\ \hline\hline
        Proposed & \underline{98.59} & \textbf{9.94} & \underline{87.24} & \textbf{6.74}\\
        w/o fine-tune & 98.34& 66.23  & 85.72 & 41.05 \\\hline
        Adversarial & \textbf{98.78} & 28.44 & \textbf{87.78} & 14.59\\
        SA-VAE~\cite{sun2018learning} & 98.44 & \underline{11.60} & 87.00 & \underline{6.82}\\ 
        Factorization~\cite{srivatsan2019deep} & N/A$^*$ & 50.77 & N/A$^*$ & 33.90\\
        FANnet~\cite{roy2020stefann} & N/A$^*$ & 76.64 & N/A$^*$ & 37.26\\\hline
  \end{tabular}
\end{table}
\subsubsection{Character recognition}\label{sec:experiment2}
We then evaluated the disentangled features in the character recognition task (26 classes). Different from the font recognition experiments in the previous section, we used the (style or content) features from all Google Fonts for training an MLP for character recognition and then use the features from Adobe font and Capitals64 for testing. MLP has the same structure as the font recognition experiment. 
\par
Table~\ref{table:nn-char} shows the test accuracy of character recognition on Adobe font and Capitals64. If higher accuracy is achieved by the content features, they are expected to represent `x'-ness (e.g., `A'-ness) more appropriately. 
Good style features, on the other hand, are expected to result in lower character recognition accuracy. \par
Similar to the font recognition experiment, the content feature extracted by the proposed method achieved the near-best accuracies (98.59\% and 87.24\%) for the datasets. The style feature by the proposed method achieved the best accuracies for the datasets. By combining the results of Tables~\ref{table:nn-font}~and~\ref{table:nn-char}, we can conclude that the proposed method can achieve the most disjoint and accurate disentanglement of font images into style and content features.\par

One important observation in Table~\ref{table:nn-char} is the large difference between the proposed method and ``w/o fine-tune'' in the style feature-based character recognition ($9.94\%$ and $66.23\%$ for Adobe and $6.74\%$ and $41.05\%$ for Capitals64). The results show that the style features before the fine-tuning step still have `x'-ness and thus the character recognition accuracy could not be minimized. Note that the content feature-based font recognition result in Table~\ref{table:nn-font} shows minimum accuracies (2.60\% and 7.32\%) even before fine-tuning -- this is because the character classifier in the pre-training model helps to exclude style information from the content feature. In contrast, because the style feature received no such treatment in the pre-training step, `x'-ness could not be excluded. 

\subsection{Performance evaluation via one-shot font generation experiment}\label{sec:font-gen}
As another evaluation of the disentangled features in an application scenario, we performed one-shot font generation, which generates all font images from `A' to `Z' by using a style feature extracted from only a single image of an arbitrary character class, such as `K.'  
In the experiment, we first choose a single font image $X_{i,j}$ from the $i$th test font (in Adobe or Capitals64) and extract its style feature $z^s_{i,j}$.
Then, we generate $\left\{\hat{X}_{i,j'} \mid \forall j'\right\}$ by using $z^s_{i,j}$ and the content features $\left\{z^c_{i',j'}\mid \forall j'\right\}$. 
A randomly selected $i' (\neq i)$th font is used for collecting the content features.
If $z^s_{i,j}$ represents the complete style information of the $i$th font and $\left\{z^c_{i',j'}\mid \forall j'\right\}$ represent general `$j'$'-ness (e.g.,  `A'-ness), it is possible to expect the near-perfect generation $\hat{X}_{i,j'} \sim X_{i,j'}$ from $z^s_{i,j}$ and $z^c_{i',j'}$, for all $j'$s.
\par
\begin{table}[t]
 \caption{Performance on one-shot font generation experiment. *In a few cases, generation is totally failed, and an empty image is given. Those cases are excluded from the evaluation because they make HD and CD an infinite value.}
 \vspace{-2mm}
 \label{table:generation-adobe}
 \centering
 \scalebox{0.9}{
  \begin{tabular}{l|r|r|r|r|r|r|r|r|r|r}
    \hline
         & \multicolumn{5}{c|}{Adobe font}&\multicolumn{5}{c}{Capitals64}\\
         \cline{2-11}
          &MSE $\downarrow$ & MAE $\downarrow$ & HD* $\downarrow$ & CD* $\downarrow$ & IoU $\uparrow$
         & MSE $\downarrow$ & MAE $\downarrow$ & HD* $\downarrow$ & CD* $\downarrow$ & IoU $\uparrow$ \\ \hline\hline
        Proposed & \underline{0.0886} & \textbf{0.1096} & \textbf{7.247} & \textbf{3.254}  & \textbf{0.6685} &\underline{0.1338} & \textbf{0.1618} & 12.586 &6.019& \underline{0.4561} \\ 
        Adversarial & 0.0964 & \underline{0.1179} & \underline{7.309} & 3.421 & \underline{0.6490} 
        &0.1398 & \underline{0.1678} & 11.835 &\underline{5.776} & 0.4489\\ 
        SA-VAE~\cite{sun2018learning}& 0.1029 & 0.1698 & 8.145&  4.255 & 0.5958 & 0.1507 &0.2278 & \underline{10.982} &5.928&0.4325\\ 
        Factorization~\cite{srivatsan2019deep}& \textbf{0.0861} & 0.1368 & 7.428 &\underline{3.397} & 0.6444 & \textbf{0.1199}& 0.1888  & \textbf{9.596} & \textbf{4.776}& \textbf{0.4781}\\
        FANnet~\cite{roy2020stefann} & 0.0983 & 0.1216 &20.051& 5.001  & 0.6217 &0.1427 &0.1713 & 23.544&9.724& 0.4000\\ \hline
  \end{tabular}}
\end{table}
\begin{figure}[t]
    \centering
    \includegraphics[width=0.9\linewidth]{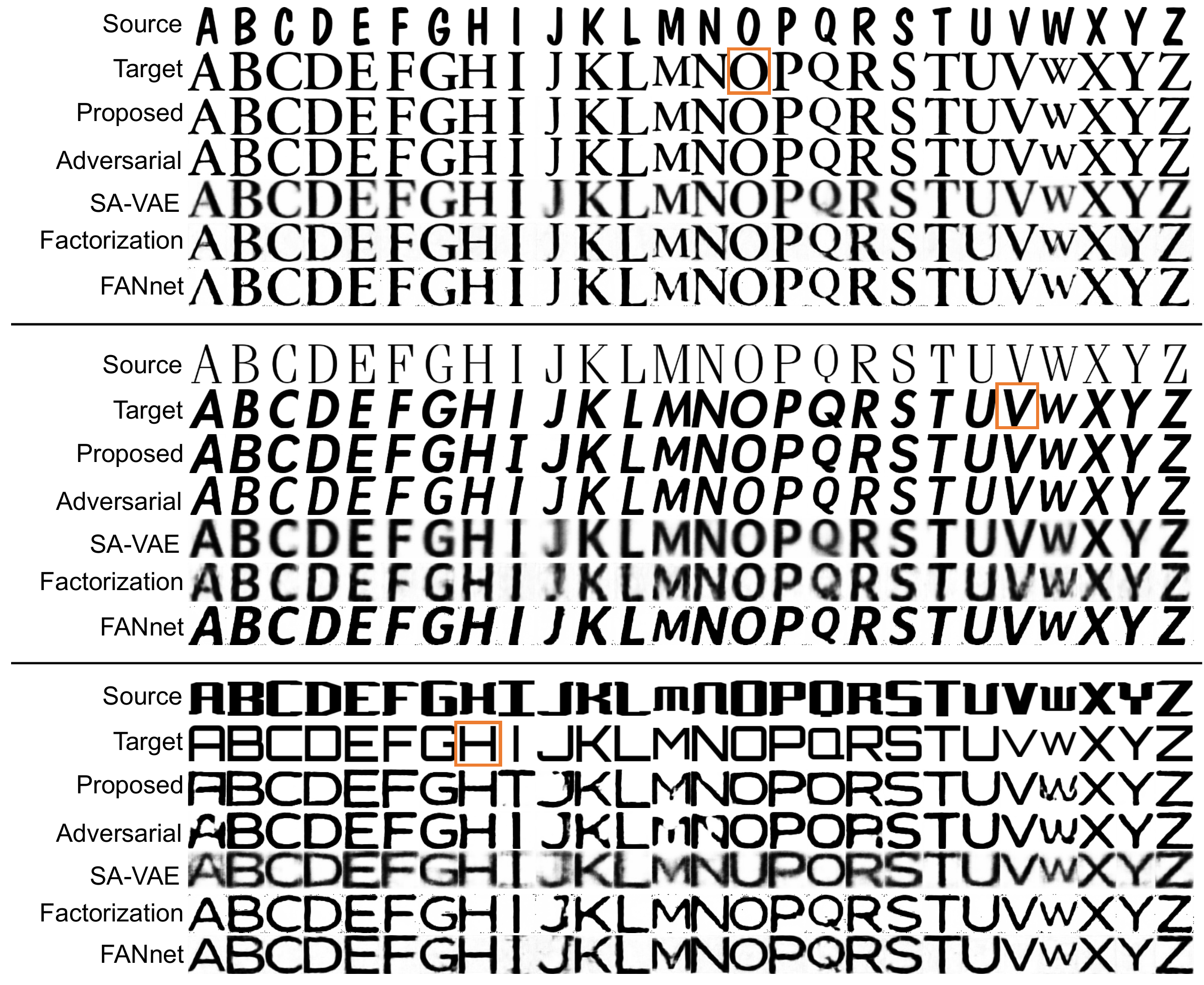}\\[-3mm]
    \caption{Examples of one-shot font generation results. The orange box shows the font used for extracting a style feature. Top and middle examples are adobe font, and the bottom font is Capitals64.}
    \label{fig:generation}
\end{figure}
We use five metrics for evaluating the quality of the generated images: Mean Square Error (MSE), Mean Absolute Error (MAE), Hausdorff distance (HD), Chamfer distance (CD), and Intersection over Union (IoU). MSE and MAE directly evaluate image differences.
HD, CD, and IoU have been used for evaluating font images frequently~\cite{kang2022shared,wang2020attribute2font,jiang2019scfont}. 
Otsu binarization and Canny edge detection are used to calculate HD and CD. After binarizing images with the Otsu method, IoU is calculated.
\par
We compare the proposed method with the comparative methods in Section~\ref{sec:comparative-methods} in this experiment. As shown in \textbf{Factorization} and \textbf{FANnet} cannot perform content disentanglement, instead use the trained parameter or content label as content input. 
Therefore, these two methods have significant advantages.
\par 
The quantitative evaluation of the font generation results is shown in Table~\ref{table:generation-adobe}. Except for the CD and HD of Capitals64, the proposed method achieves the best or second-best results in all metrics. 
Capitals64 includes a variety of fancy fonts with reduced readability; therefore, the proposed method occasionally makes mistakes in extracting content features. 
In contrast, Factorization can ignore such difficulties and generate font images with precise content because of using trained parameters as content features.
It is worth noting that the proposed method outperforms or equivalent the methods that employ the content label or ``trained'' content feature such as Factorizaiton, despite these disadvantages.
As noted above, they are free of character-class misrecognition during content disentanglement. The fact that the proposed method still outperforms them demonstrates the high quality of our disentangled features.
\par

Fig.~\ref{fig:generation} shows examples of font generation. ``Target'' is the font to be completed, and only the single character image with the orange bounding box is given as $X_{i,j}$ to extract the style feature $z^s_{i,j}$. The font of ``Source'' is the randomly selected test font $i'$ for extracting the content feature $z^c_{i',j}$.
As shown in the top examples, the proposed method could extract the serif font of target `O' and generate the other classes by using the style feature. Notably, the proposed method can generate the detailed serif parts more precisely than others.
The middle example demonstrates that the target `V' has slight oblique styles, and the proposed method capture such slight style information. Several methods do not capture the slight style.
It is interesting to note that the style is appropriately extracted from a curve-only letter `O' and a straight-line-only letter `V.'
The bottom example show not simple serif or sans-serif font but a little bit fancy style font. Notably, only the proposed method can generate several letters, such as `A', in a square-shaped style that matches the target font.

\par
%


\section{Discussion}\label{sec:discussion}
\subsection{What is `A'?}
The question ``What is `A'?'' is an essential problem of epistemology and the motivation of Hofstadter's ``vertical and horizontal problem''~\cite{metamagical1985}. It is also a critical question in pattern recognition in computer science because it is related to the question, ``What is the ground truth?'' Fig.~\ref{fig:matrix-pca}~(a) and other results support that our total disentanglement method can give almost the same content feature $z^c_{i,j}$ from all images of the character class $j$ after removing the style feature $z^s_{i,j}$. Consequently, since $j=1$ corresponds to the class `A', $z^c_{*, 1}$ will be an answer to the question and be the `A'-ness~\cite{metamagical1985}~\footnote{If we want  a unique `A'-ness vector rather than a slightly different vector set $\{z^c_{i, 1}\mid \forall i\}$, we can take the average of $z^c_{i, 1}.$}.\par
Interestingly, we cannot picture a content feature $z^c_{*, 1}$ as an image of `A' by itself.  We can do it only by combining it with a style feature. This reminds us of Plato's {\em allegory of the cave}, which he uses to explain the theory of {\em ideas} (, or the theory of Forms). According to the allegory, we cannot see the idea by itself but can see it as a shadow by projecting light onto it. A content feature, a style feature, and a font image correspond to an idea, light, and shadow in our scenario. `A'-ness is invisible and only becomes visible by attaching a style. Since different lights make different shadows, different styles make different images. 

\subsection{Limitation}
\begin{figure}[t]
    \centering
    \includegraphics[width=0.89\linewidth]{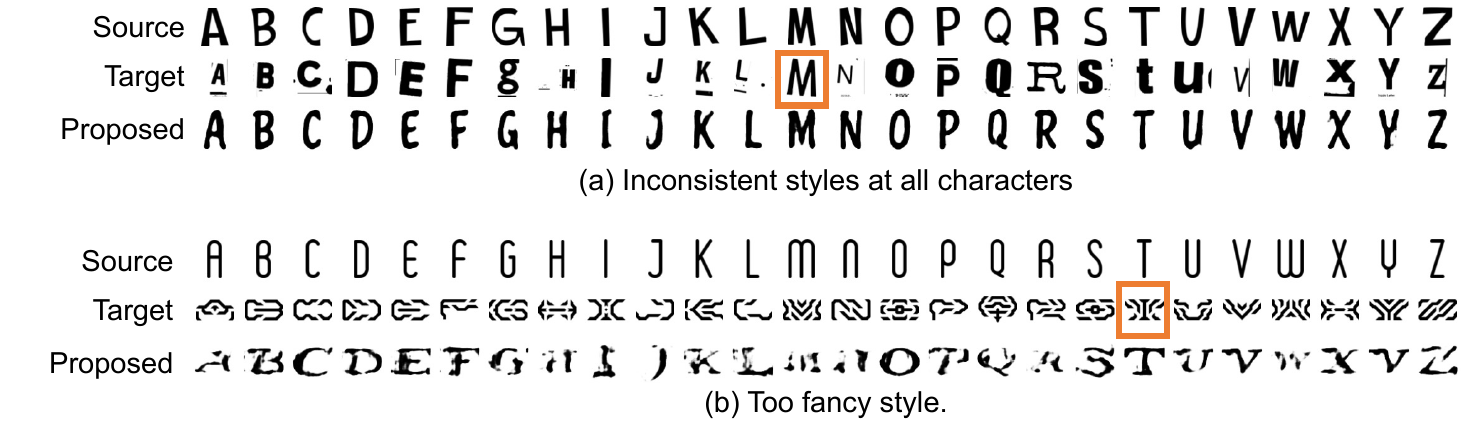}\\[-4mm]
 \caption{Failures in the one-shot generation result from the proposed method.}
    \label{fig:limitation}
\end{figure}

Fig.~\ref{fig:limitation} shows two failure cases of the one-shot font generation by the proposed method. Because our method assumes that font images in a font set will have the same style, it cannot deal with the font of (a), which has different font styles for each character. Case (b) shows a failure for a different reason; the target font has a very unusual font style, and it was difficult to generate them with an appropriate style because of the decoder performance.
In a future work, we will integrate font style features extracted by total-disentangle to diffusion model for generating more complicated fonts.
\par
In the other limitation, since our current formulation deals with bitmap images as input and output, the output font image frequently has blurry and noisy strokes. 
Recently, several font analysis methods utilize an implicit representation~\cite{reddy2021multi,liu2022learning} or an outline representation~\cite{wang2021deepvecfont,lopes2019learned,wang2023deepvecfont}.
We can modify our method to deal with such representations for a sharper stroke representation.


\section{Conclusion}
We proposed a total disentanglement method for font images by fully utilizing the characteristics of font images, which have two independent features: character class and font style. By introducing variance losses and a careful pre-training step, the proposed model can decompose an input font image into its content and style features, while satisfying the consistency conditions that all images of the font $j$ have almost the same style features and all images of the character class $j$ have almost the same content features.\par
Several quantitative and qualitative experiments confirmed that the proposed method performs as expected. First, visualizations of disentangled features demonstrate that the style and content features surely satisfy the aforementioned consistent conditions. This means that ``the vertical and horizontal problems'' by Hofstadter in 1985 are solvable. Second, all of the evaluations on font classification, character classification, and one-shot font generation tasks 
show that the proposed method outperforms other methods that deal with font styles; these results also support the high quality of our disentangled features. \par
Last but not least, the proposed disentanglement method answers
an open and general problem of pattern recognition research, namely, ``what is `A'?''  with recent machine learning technologies (The idea behind the vertical and horizontal problems is to answer this question.). We hope that our results will contribute to a deeper understanding of pattern recognition from a viewpoint of cognitive science, especially, epistemology.

\vspace{-4mm}
\subsubsection{\ackname} This work was supported in part by JSPS KAKENHI JP22H00540 and JP25H01149, and JST CRONOS-JPMJCS24K4.

%
%
%
\bibliographystyle{splncs04}
%
\bibliography{ref}




\end{document}